\documentclass[12pt,a4paper]{cibb}

\makeatletter
\providecommand{\@ordinalM}[2]{#1}
\makeatother

\usepackage{subfigure,graphicx}
\usepackage{amsmath,amsfonts,latexsym,amssymb,euscript,xr}
\usepackage{booktabs}
\usepackage[nodayofweek]{datetime}
\usepackage{hyperref}
\usepackage{fmtcount}
\usepackage[english]{datenumber}
\usepackage[absolute]{textpos}

\usepackage[table]{xcolor}
\usepackage{color,colortbl,tabularx}

\usepackage[english]{babel}
\usepackage[protrusion=true,expansion=true]{microtype}
\usepackage{amsmath,amsfonts,amsthm}
\usepackage{pifont}

\definecolor{LightBlue}{rgb}{0.88,0.9,0.9}

\title{\Large $\ $\\ \bf ClinAgent: A ReAct-Based Agent for Conversational Access to Clinical Trial Information}

\author{\large Antonino Vaccarella$^{1,2,3}$, Riccardo Cantini$^{*,2}$, Domenico Talia$^{2}$, Paolo Trunfio$^{2}$, Marianna Talia$^{4}$, Rosamaria Lappano$^{4,5}$, and Marcello Maggiolini$^{4}$}
\address{\footnotesize $\ $\\$^1$ Department of Computer Science, University of Pisa, Pisa, Italy \\
$^2$ Department of Computer Engineering, Modeling, Electronics and Systems, University of Calabria, Rende, Italy.\\
$^3$ Institute of Information Science and Technologies ``Alessandro Faedo'', National Research Council, Pisa, Italy.\\
$^4$ Department of Pharmacy, Health and Nutritional Sciences, University of Calabria, Rende, Italy. \\
$^5$ Department of Experimental and Clinical Medicine, University ``Magna Gr{\ae}cia'' of Catanzaro, Catanzaro, Italy. \\
\bigskip
ORCID codes: AV 0009-0004-0428-540X; RC 0000-0003-3053-6132; DT 0000-0003-1910-9236; PT 0000-0002-5076-6544; MT 0000-0002-3440-5618; RL 0000-0002-9374-9701; MM 0000-0002-7485-854X.
\bigskip
\newline
$^*$corresponding author: rcantini@dimes.unical.it
}

\abstract{\small Agentic AI, Retrieval-Augmented Generation, Clinical Trials, ReAct, Medical Informatics. \normalsize
\\[17pt]
{\bf Abstract.}
Querying clinical trial registries remains a manual and error-prone process, requiring researchers to navigate large volumes of semi-structured data without support for natural-language interaction or cross-source synthesis. To address this, we introduce \textit{ClinAgent}, a conversational system based on agentic Retrieval-Augmented Generation (RAG) that enables clinicians and researchers to query clinical trial information in plain language and receive grounded, up-to-date responses across multi-turn interactions. 
The system centers on a Large Language Model (LLM) agent following the ReAct paradigm, which iteratively reasons over queries, selects among a set of integrated tools, and refines its actions based on intermediate outputs. These tools include a ClinicalTrials.gov search interface, a PubMed module, and a Python-based analyzer operating on a locally cached structured dataset of clinical trials. 
We evaluate the system using a three-phase framework assessing operational effectiveness, planning quality, tool-use efficiency, and expert qualitative judgments, comparing three LLM backends: Gemini 3.0 Flash and two variants of DeepSeek V3.2 (thinking and non-thinking). Results reveal complementary strengths, with DeepSeek (thinking mode) excelling in planning quality, while Gemini achieves the highest overall performance and strongest expert ratings. 
Overall, our findings highlight the potential of agentic AI systems to improve the accessibility and synthesis of clinical trial information, supporting more efficient and user-centered biomedical research workflows.
}

\begin{document}

\renewcommand{\thefootnote}{}
\footnotetext{\small{Article version: \datedate $\;$ h\currenttime  $\;$ CET}}

\thispagestyle{myheadings}
\pagestyle{myheadings}
\markright{\tt Proceedings of CIBB 2026}%check year

\section{Introduction}

Clinical trial registries, such as ClinicalTrials.gov, index over a million registered studies spanning several therapeutic areas and study phases. Despite the scale of this resource, querying it remains a largely manual activity, with users relying on keyword-based search through web interfaces or bulk CSV exports inspection with general-purpose tools~\cite{paunic2025cross,tse2018avoid}. Neither approach supports natural language interaction nor offers any mechanism for synthesising information across multiple records or contextualising registry data against the broader scientific literature.

Large Language Models (LLMs) have changed what is technically feasible here, yet their tendency to hallucinate --- generating plausible but factually incorrect content --- is not acceptable in a clinical context. Retrieval-Augmented Generation (RAG) addresses this by grounding model responses in documents retrieved at query time from verified external sources, substantially reducing hallucination without requiring model retraining. Standard RAG pipelines, however, operate in a single \textit{retrieval-then-generate} pass, which limits their ability to handle queries that require gathering information from multiple sources or that benefit from iterative refinement. Agentic approaches address this through an explicit reasoning loop: under the ReAct paradigm~\cite{yao2022react}, an agent alternates between formulating a reasoning step, invoking a tool, and observing the result, repeating the cycle until sufficient evidence has been accumulated.

Building upon these considerations, this paper makes two contributions. First, we introduce \textit{ClinAgent}, a conversational AI system built on agentic RAG that enables clinicians and researchers to query clinical trial information in plain language, integrating three specialised tools under a single LLM agent. These tools comprise a \textit{ClinicalTrials.gov search} module for retrieving trial records, a \textit{PubMed interface} for accessing and summarising relevant biomedical literature, and a \textit{Python-based analysis} component that operates over a locally cached, structured dataset of clinical trials accumulated during user interactions.
Second, we propose a three-phase evaluation framework covering $(i)$ operational effectiveness, $(ii)$ plan quality and tool-use efficiency, and $(iii)$ expert qualitative assessment, and apply it comparatively across three recent LLM backends---Gemini 3.0 Flash and two versions of DeepSeek V3.2, i.e., the standard model and a reasoning-enabled variant (thinking mode). Results reveal non-trivial trade-offs across backends: DeepSeek leads on plan quality in thinking mode, consistent with its explicit reasoning capabilities, while Gemini 3.0 Flash achieves the highest overall score and strongest expert ratings. These findings suggest that agentic RAG is a practically viable approach to clinical trial information retrieval, and that the trade-off between reasoning depth and factual reliability is a critical consideration for backend selection in clinical deployment.

\section{Background and Related Work}
Generative language models, despite their fluency, are prone to factual errors and cannot incorporate knowledge postdating their training. Retrieval-Augmented Generation (RAG) addresses both issues by grounding outputs in dynamically retrieved external documents, thereby improving factual accuracy and enabling access to up-to-date information without retraining. Yang et al.~\cite{yang2025retrieval} systematically analyze RAG in medical AI, showing that grounding responses in curated knowledge reduces hallucinations and mitigates pre-training biases, making RAG well-suited to high-stakes domains. Miao et al.~\cite{miao2024integrating} apply these principles in nephrology, demonstrating that standard LLMs without retrieval exhibit hallucination rates too high for clinical use, thereby establishing domain-specific grounding as an operational requirement rather than a theoretical refinement. OpenEvidence~\cite{openevidence} is a clinical question-answering system that provides grounded answers to medical queries by retrieving evidence from PubMed and curated clinical guidelines. While it demonstrates the viability of AI-assisted medical QA in high-stakes settings, it operates over general clinical knowledge rather than trial registries, and does not support structured registry querying, programmatic analysis of trial records, or agentic multi-step reasoning across heterogeneous tools and knowledge sources.

The system most closely related to \textit{ClinAgent} is TrialGPT~\cite{jin2024matching}, which uses LLMs to match individual patients to eligible clinical trials by evaluating eligibility criteria and aggregating judgments into a trial-level score, achieving expert-level accuracy and significantly reducing manual matching time. While TrialGPT and \textit{ClinAgent} operate in the same domain, they address different problems: TrialGPT focuses on patient recruitment, whereas \textit{ClinAgent} supports open-ended natural-language queries by synthesizing grounded responses from heterogeneous sources. To the best of our knowledge, no existing system combines agentic RAG with multi-source retrieval across both a trial registry and biomedical literature for conversational clinical information access, a capability with direct practical relevance for clinicians and researchers who require timely, grounded access to trial evidence without the overhead of manual registry navigation and inspection.

\section{Proposed System}

\begin{figure}[h]
    \centering
    \includegraphics[width=0.95\linewidth]{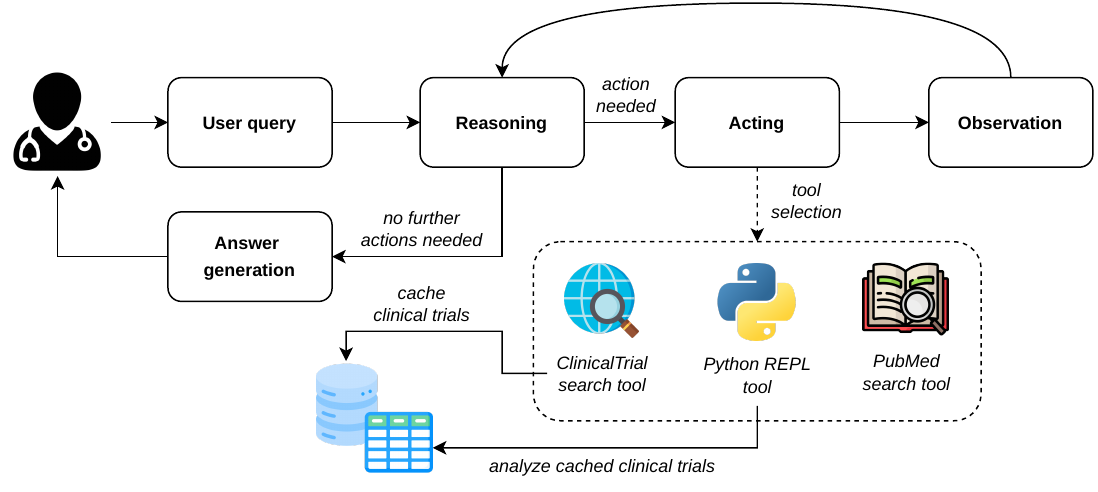}
    \caption{High-level architecture of \textit{ClinAgent}. The LLM agent 
    operates a closed reasoning--action--observation loop, selecting among three 
    specialised tools within a multi-turn user interaction.}
    \label{fig:architecture}
\end{figure}

\textit{ClinAgent} is an agentic AI conversational system for clinical trial information retrieval built on the ReAct paradigm~\cite{yao2022react}. Upon receiving a natural language query, the LLM agent does not attempt to answer immediately; instead, it engages in an iterative \textit{reasoning--acting--observation} loop. In the \textit{reasoning} step, the agent interprets the query, decomposes it into sub-goals, and formulates a retrieval plan. In the \textit{acting} step, it invokes one of the available tools, delegating retrieval or computation to the appropriate external component. In the \textit{observation} step, it examines the tool's output and decides 
whether the accumulated evidence is sufficient to produce a response or whether further actions are needed. The overall architecture is shown in Figure~\ref{fig:architecture}.

The system integrates three specialised tools. The \textit{ClinicalTrial search} tool wraps a module that translates free-text queries into well-formed requests to the ClinicalTrials.gov REST API. An LLM first extracts structured parameters from the query (e.g., medical condition, intervention, NCT identifier, and 
result-availability filter), which are validated against a Pydantic schema before being serialised into a targeted HTTP request. Retrieved records are cached locally in a CSV dataset, ensuring that data fetched during a session remains available for subsequent queries without redundant API calls. The \textit{Python Read-Eval-Print Loop (REPL)} tool allows the agent to perform programmatic analyses on this locally cached dataset using Pandas-based code, enabling filtering, aggregation, and statistical operations directly over retrieved trial records. The \textit{PubMed search} tool accesses the NCBI Entrez API via LangChain, enabling the agent to retrieve relevant biomedical literature and ground its answers in peer-reviewed evidence as well as trial data.

The agent dynamically selects tools based on the query and observed results, allowing high autonomy and resilience. If a primary tool fails or returns insufficient information, the agent replans and attempts an alternative strategy without external intervention, ensuring seamless user interaction. This iterative, self-correcting approach enhances reliability and enables adaptation to complex, evolving queries, making the agent well-suited for high-stakes biomedical applications where precision and evidence-based reasoning are essential.

The workflow is implemented using \textit{LangChain} and \textit{LangGraph} frameworks, the latter providing native support for the cyclic execution graphs required by agentic architectures. Conversation state is persisted across successive requests via a \textit{MemorySaver} component, enabling the agent to resolve follow-up queries by reference to earlier exchanges, thus allowing effective multi-turn interaction with the user.

\section{Experimental Evaluation}

We evaluate \textit{ClinAgent} using a three-phase framework designed to assess distinct aspects of system performance: \textit{operational effectiveness} (Phase~1), \textit{plan quality} and \textit{tool-use efficiency} (Phase~2), and \textit{qualitative assessment} by clinical domain experts (Phase~3). Three LLM backends were included: Gemini 3.0 Flash and two variants of DeepSeek V3.2 (thinking and non-thinking mode). Each model was evaluated on a standardised query set comprising factual retrieval of trial attributes, comparative analysis across studies, literature-based questions, and mixed queries combining trial data with published evidence from PubMed. The full results across all phases are reported in Table~\ref{tab:results} and discussed in the following sections.

\begin{table}[htb!] \small
\centering
\caption{Summary of results across all three evaluation phases. For each metric, the best result across backends is highlighted in bold. DeepSeek's variant annotated with \textsc{tm} was used in thinking mode.}
\label{tab:results}
    \begin{tabularx}{\textwidth}{ X l c c c }
        \toprule
        \textbf{Evaluation Phase} & \textbf{Metric} & \textbf{Gemini 3.0 Flash} & \textbf{DeepSeek V3.2 (\textsc{tm})} & \textbf{DeepSeek V3.2} \\
        \midrule
        \rowcolor{LightBlue} Phase 1 & Goal Achievement Rate & \textbf{1.00} & 0.94 & 0.88 \\
        \midrule
        \rowcolor{LightBlue} Phase 2 & ReAct Quality & \textbf{0.75} & 0.73 & 0.62 \\
                                     & \quad Plan Quality & 0.94 & \textbf{0.97} & 0.85 \\
        \rowcolor{LightBlue}         & \quad Tool Use Efficiency & \textbf{0.56} & 0.49 & 0.39 \\
        \midrule
        \rowcolor{LightBlue} Phase 3 & Aggregate Expert Score & \textbf{0.66} & 0.45 & 0.65 \\
                                     & \quad Clarity & 0.73 & 0.63 & \textbf{0.75} \\
        \rowcolor{LightBlue}         & \quad Completeness & 0.58 & 0.35 & \textbf{0.68} \\
                                     & \quad Logical consistency & \textbf{0.75} & 0.35 & 0.65 \\
        \rowcolor{LightBlue}         & \quad Correctness & \textbf{0.68} & 0.33 & \textbf{0.68} \\
                                     & \quad Factuality & \textbf{0.65} & 0.28 & \textbf{0.65} \\
        \rowcolor{LightBlue}         & \quad Relevance to question & \textbf{0.65} & 0.60 & 0.45 \\
                                     & \quad Quality of interaction & 0.68 & \textbf{0.75} & \textbf{0.75} \\
        \rowcolor{LightBlue}         & \quad Clinical relevance & \textbf{0.65} & 0.30 & 0.60 \\
        \midrule
                                     & Overall score & \textbf{0.80} & 0.71 & 0.72 \\
        \bottomrule
    \end{tabularx}
\end{table}

\subsection{Phase 1 --- Operational Effectiveness}

Phase~1 establishes whether each candidate LLM reliably completes a given set of queries $Q$. For each query $q \in Q$, a binary \textit{success} function is defined as follows:
\begin{equation}
    \text{success}(q) =
    \begin{cases}
        1 & \text{if the agent produces a response relevant to the query} \\
        0 & \text{otherwise}
    \end{cases}
\end{equation}
The \textit{Goal Achievement Rate} (GAR) is then computed as the fraction of successful queries:
\begin{equation}
    \text{GAR} = \frac{1}{|Q|} \sum_{q \in Q} \text{success}(q)
\end{equation}
Gemini 3.0 Flash achieved a GAR of $1.00$, successfully handling all queries in the evaluation set. DeepSeek V3.2 also demonstrated strong operational reliability, with performance varying across modes: the reasoning-enabled (thinking) mode attains a higher GAR ($0.94$) than the non-thinking mode ($0.88$). The observed failures are primarily attributable to incomplete tool invocations or premature termination of the reasoning loop before sufficient evidence is gathered.

\subsection{Phase 2 --- Plan Quality and Tool-Use Efficiency}

Phase~2 analyzes not only whether the system reaches an answer, but also how it does so. We first define the \textit{success} function for a single action $a$ recursively to account for self-correction:
\begin{equation}
    \text{success}(a) =
    \begin{cases}
        1 & \text{if } \bigl[\top(a) \wedge \neg\,\text{redundant}(a)\bigr]\, \vee \, \bigl[\bot(a) \wedge \text{success}(\text{replan}(a))\bigr] \\
        0 & \text{otherwise}
    \end{cases}
\end{equation}
where $\top(a)$ denotes that the invocation of action $a$ succeeds, $\bot(a)$ that it fails, $\text{redundant}(a)$ that $a$ retrieves information already present in the current context, and $\text{replan}(a)$ the corrective action generated by the agent following the failure of $a$. In this definition, an action that fails but is followed by a successful replan is credited, while one that fails irrecoverably or duplicates already-retrieved information is penalized.
We then define the \textit{Plan Quality} (PQ) and \textit{Tool Use Efficiency} (T$_{\text{eff}}$) metrics. PQ measures the proportion of actions in the agent's plan that are both successful and non-redundant, crediting self-correction through replanning. 
T$_{\text{eff}}$ measures the fraction of tool invocations that succeed on the first attempt, without considering recovery, thus penalising agents that reach a correct answer only after multiple failed invocations:
\begin{equation}
    \text{PQ} = \frac{1}{|Q|} \sum_{q \in Q} \frac{1}{|A_q|} 
    \sum_{a \in A_q} \text{success}(a)\,, \qquad
    \text{T$_{\text{eff}}$} = \frac{1}{|Q|} \sum_{q \in Q} \frac{1}{|A_q|} 
    \sum_{a \in A_q} \mathbf{1}[\top(a)]
\end{equation}
Here, $A_q$ is the set of actions executed by the agent for query $q$, and $\mathbf{1}[\top(a)]$ is the indicator function that equals $1$ if action $a$ succeeds on the first attempt and $0$ otherwise. A composite \textit{ReAct Quality} score is finally defined as the average of PQ and T$_{\text{eff}}$, rewarding both robust multi-step planning and resilience to failure, as well as precision and efficiency in tool selection.
DeepSeek V3.2 in thinking mode achieved the highest PQ ($0.97$), consistent with its explicit reasoning capabilities, which support more structured multi-step planning. Conversely, Gemini 3.0 Flash attains the highest T$_{\text{eff}}$ ($0.56$), reflecting more reliable first-attempt tool selection and invocation. On the composite ReAct Quality score, Gemini 3.0 Flash ($0.75$) marginally outperformed reasoning-enabled DeepSeek V3.2 ($0.73$), while DeepSeek V3.2 in non-thinking mode ($0.62$) lags substantially behind.

\subsection{Phase 3 --- Expert Qualitative Assessment}

In this phase, experts evaluated system responses across four test scenarios---factual retrieval, comparative analysis, literature-based questions, and mixed queries---using a five-point Likert scale over eight criteria: \textit{clarity}, \textit{completeness}, \textit{logical consistency}, \textit{correctness}, \textit{factuality}, \textit{relevance to the question}, \textit{quality of interaction}, and \textit{clinical relevance}. Scores were normalized to the $[0, 1]$ interval. 
Gemini 3.0 Flash exhibits the most balanced profile, achieving the highest scores in logical consistency and clarity, with comparatively lower performance in completeness, suggesting well-structured and accurate responses that occasionally omit relevant aspects. DeepSeek V3.2 in non-thinking mode performs strongly in clarity and completeness but drops markedly in relevance, indicating a tendency toward detailed yet less focused answers. In contrast, the reasoning-enabled version of DeepSeek V3.2 scores substantially lower than both models in factuality, correctness, and clinical relevance, despite leading in interaction quality.

\subsection{Overall Results}

The \textit{overall score} is computed as the average of Phase~1, Phase~2, and Phase~3 assessment scores, aggregated across experts and metrics. Gemini 3.0 Flash achieves the highest overall score ($0.80$), emerging as the most balanced backend for deployment in this setting. DeepSeek V3.2 variants perform comparably at the aggregate level, but exhibit complementary strengths: the standard variant ($0.72$) yields clearer and more complete responses according to expert evaluations, whereas the reasoning-enabled variant ($0.71$) demonstrates superior multi-step planning and greater resilience to tool failures, albeit at the cost of reduced factual reliability.

\section{Conclusion}

We presented \textit{ClinAgent}, a ReAct-based agentic AI system for conversational clinical trial information retrieval, along with a three-phase evaluation framework to assess agentic approaches in biomedical information access. Evaluation across three different LLMs highlights non-trivial trade-offs in backend selection: Gemini 3.0 Flash achieves the highest overall score and the strongest expert ratings, making it the recommended choice for deployment, whereas DeepSeek's reasoning-enabled variant offers measurable advantages in multi-step planning at the cost of lower factual reliability. Future work will focus on extending \textit{ClinAgent} to include additional domain-specific sources, such as regulatory databases and institutional electronic health records.

\section*{Conflict of interests}
\label{sec:CONFLICT-OF-INTERESTS}
The authors declare no conflicts of interest.

\section*{Funding}
\label{sec:FUNDING}
We acknowledge financial support from “FAIR – Future Artificial Intelligence Research” project - CUP H23C22000860006, and “ECHO-TWIN -- Edge-Cloud-HPC Optimized Twins” project, an initiative of the National Center ICSC-HPC, Big Data and Quantum Computing. 

\section*{Availability of data and software code}
\label{sec:AVAILABILITY}
The developed \textit{ClinAgent} system is publicly available as open source at the following URL: \small{\url{https://github.com/antoninovaccarella/clinical-chat}}.

\footnotesize
\bibliographystyle{unsrt}
\bibliography{bibliography_CIBB_file.bib} 
\normalsize

\end{document}